\documentclass[sn-mathphys]{sn-jnl}% Math and Physical Sciences Reference Style
\jyear{2021}%
\theoremstyle{thmstyleone}%
\theoremstyle{thmstyletwo}%
\theoremstyle{thmstylethree}%
\usepackage{amsmath}
\usepackage{caption}
\usepackage{subcaption}
\usepackage{textalpha}
\usepackage{dsfont}
\usepackage{amsmath}
\usepackage{float}
\usepackage{multirow}
\usepackage{hhline}
\usepackage{xcolor,pifont}
\usepackage{amssymb}% http://ctan.org/pkg/amssymb
\usepackage{pifont}% http://ctan.org/pkg/pifont
\newcommand{\xmark}{\ding{55}}%
\newcommand*\colourcheck[1]{%
  \expandafter\newcommand\csname #1check\endcsname{\textcolor{#1}{\ding{52}}}}
\colourcheck{blue}
\colourcheck{green}
\colourcheck{red}
\usepackage{booktabs}
\usepackage{multirow}
\usepackage{hhline}
\begin{document}
\title[Bridging the Performance Gap between DETR and R-CNN]{Bridging the Performance Gap between DETR and R-CNN for Graphical Object Detection in Document Images.}
\newcommand{\orcidauthorA}{0000-0003-0456-6493} % Add \orcidA{} behind the author's name
\newcommand{\orcidauthorB}{0000-0002-0536-6867} % Add \orcidB{} behind
\newcommand{\orcidauthorC}{0000-0002-7052-979X}

\author*[1,2,3]{\fnm{Tahira} \sur{Shehzadi}}\email{tahira.shehzadi@dfki.de}

\author[1,2,3]{\fnm{Khurram Azeem} \sur{Hashmi}}
\author[1,2,3]{\fnm{Didier} \sur{Stricker}}
\author[4]{\fnm{Marcus} \sur{Liwicki}}
\author[1,2,3]{\fnm{Muhammad Zeshan} \sur{Afzal}}

\affil[1]{\orgdiv{Department of Computer Science}, \orgname{Technical University of Kaiserslautern}, \postcode{67663}, \country{Germany}}
\affil[2]{\orgname{Mindgarage Lab}, \city{Kaiserslautern}, \postcode{67663}, \country{Germany}}

\affil[3]{\orgdiv{Augmented Vision}, \orgname{German Research Institute for Artificial Intelligence (DFKI)}, \city{Kaiserslautern}, \postcode{67663}, \country{Germany}}

\affil[4]{\orgdiv{Department of Computer Science}, \orgname{Luleå University of Technology},\city{Luleå}, \postcode{97187}, \country{Sweden}}

\abstract{This paper takes an important step in bridging the performance gap between DETR and R-CNN for graphical object detection. Existing graphical object detection approaches have enjoyed recent enhancements in CNN-based object detection methods, achieving remarkable progress. Recently, Transformer-based detectors have considerably boosted the generic object detection performance, eliminating the need for hand-crafted features or post-processing steps such as Non-Maximum Suppression (NMS) using object queries. However, the effectiveness of such enhanced transformer-based detection algorithms has yet to be verified for the problem of graphical object detection. Essentially, inspired by the latest advancements in the DETR, we employ the existing detection transformer with few modifications for graphical object detection. We modify object queries in different ways, using points, anchor boxes and adding positive and negative noise to the anchors to boost performance. These modifications allow for better handling of objects with varying sizes and aspect ratios, more robustness to small variations in object positions and sizes, and improved image discrimination between objects and non-objects. We evaluate our approach on the four graphical datasets: PubTables, TableBank, NTable and PubLaynet. Upon integrating query modifications in the DETR, we outperform prior works and achieve new state-of-the-art results with the mAP of 96.9\%, 95.7\% and 99.3\% on TableBank, PubLaynet, PubTables, respectively. The results from extensive ablations show that transformer-based methods are more effective for document analysis analogous to other applications. We hope this study draws more attention to the research of using detection transformers in document image analysis.}

%%================================%%
%% Sample for structured abstract %%
%%================================%%




\maketitle

\section{Introduction}
\label{sec:introduction}
Over the past few decades, the rapid growth in digital transformation in different forms like web pages, scientific publications, invoices, and financial statements has increased the storage and production of digital documents \cite{digital10}. Digitization and information extraction from such a large collection of documents is impossible for humans. So, many researchers are attracted to automatic information extraction from document images. These documents contain text and graphical page objects such as formulas, figures and tables. Even though modern Optical Character Recognition (OCR)  networks \cite{ocr8, ocr50, ocr45} can extract text information, they fail to interpret graphical page objects. The graphical object detection task in document analysis is challenging as the documents may have high variability in layout, complex backgrounds, small object size, similarity with text, and limited training data. Thus, it is essential to have an accurate graphical object detection system for document analysis.\\

For the graphical object detection task, previous methods used rule-based approaches \cite{Anh56, Gatos876, Harit89, Yalin78} and, then, CNN-based object detectors such as R-CNN \cite{Yi77, Oliveira565}, Faster R- CNN \cite{FasterTD}, and Cascade Mask R-CNN \cite{Agarwal52} have been presented. Thus, the improvement in object detection networks is directly reflected in state-of-the-art graphical object detection systems.  Earlier, people used the classical detectors in the object detection domain and transformer-based networks in the sequence prediction domain. Recently, Carion et al. \cite{att75} proposed a transformer-based architecture for object detection that performs better than all CNN-based object detectors. For the first time, Smock et al. \cite {pubtables5} used a simple DEtection TRansformer (DETR) framework for table detection and structure recognition tasks on the PubTables \cite{pubtables5} dataset and observed excellent results. However, rapid progress in transformers still need to be employed in document analysis. \\

Transformer-based object detectors \cite{shehzadi2dreview} remove the need for hand-designed components like non-maximum suppression (NMS) and anchor design used in CNN-based object detectors by introducing the concept of object queries. Object queries are learned vectors that are used to attend to all the objects in the image simultaneously and predict their class and location.
In contrast to CNN-based object detectors, which generate a fixed number of proposals or anchors that are then classified and refined, transformer-based detectors use an attention mechanism to dynamically attend to relevant parts of the image and predict the objects. This allows them to detect objects of different sizes and aspect ratios without the need for hand-designed anchor boxes.
Additionally, because transformer-based detectors do not rely on a fixed set of proposals, they can avoid the need for post-processing steps such as NMS. This simplifies the training and inference pipelines and can lead to faster and more accurate object detection.

Modifying object queries with points, anchor boxes, positive noise, and negative noise can improve the performance of graphical page object detection by providing additional localization cues, handling non-standard aspect ratios more effectively, and better differentiating between foreground and background graphical objects. These modifications can also make training more efficient and improve the generalization capabilities of the detector. By incorporating these techniques, graphical object detection models can achieve higher accuracy and efficiency, while reducing false positives and improving the overall detection performance. This paper helps in bridging the performance gap between DETR and R-CNN for graphical object detection in document images. We use the potential of the detection transformer with improved object queries. Additionally, two different pre-processing approaches are applied to the training process, further boosting the table detection performance and helping it learn more effectively. We evaluate transformer-based detectors on four popular graphical object detection datasets: TableBank \cite{tablebank8}, Publaynet \cite{PubLayNet3}, NTables \cite{NTables} and PubTables \cite{pubtables5} and compare their performance with CNN-based detectors \cite{faster23, fast15, mask86, ssd23, retinaNet68, yolox34}. \par
\begin{itemize}
 \item We present an end-to-end trainable graphical object detection framework that operates on a DEtection TRansformer (DETR) equipped with improved object queries.
  \item To the best of our knowledge, for the first time, extensive experiments are conducted on transformers for the graphical object detection domain and leverage the potential of detection transformers for this task.
  \item  We use different pre-processing techniques that boost the graphical object detection performance.  
  \item We accomplish state-of-the-art performance of transformer-based detectors on four publicly available graphical object detection datasets in scanned document images and camera-captured images.
\end{itemize}
%%%%%%%%%%%%%%%%%%%%%%%%%%%%%%%%%%%%%%%%%%
The remaining part of this paper is arranged as follows. Section \ref{sec:related_work} discusses the previous work on graphical object detection using popular computer vision approaches. Section \ref{sec:method} explains transformer-based detector and its submodules. Section \ref{sec:experiments_results} discusses the implementation details and performance analysis. Section \ref{sec:conclusion} concludes the paper and discusses future directions.

\section{Related work}
\label{sec:related_work}
\subsection{Traditional Approaches}
Previous methods used rule-based approaches for graphical objects such as table detection problems in the document analysis domain. These approaches locate tables in predefined scenarios using different rules like vertical and horizontal lines \cite{Anh56, Gatos876}, text layouts \cite{Kienin67}, keywords \cite{Harit89}, or formal patterns \cite{Yalin78}. These approaches need manual work for tuning hyper-parameters and designing rules. Many machine learning approaches \cite{Cesar38, Silva89} are proposed to minimize these heuristics dependencies. These machine learning-based approaches are explained briefly in \cite{Costa78}. Even though these approaches boost table detection performance, they use handcrafted features, Thus these methods cannot be generalized.   
\subsection{Deep Learning Based Approaches}
With the increasing progress in deep learning, many CNN-based table detection approaches have been presented that significantly improve performance. The researchers classified these approaches into semantic segmentation, bottom-up and object detection approaches.\\  
\noindent\textbf {Semantic Segmentation Based Approaches.}~
Graphical object detection such as table detection is also considered a segmentation problem, and applied the segmentation mask using current semantic segmentation networks like Fully-Convolutional Networks (FCN) \cite{fcn65} to detect tables at the pixel level in \cite{Xiao17, He761, Isaak36}. In \cite{Xiao17}, the author designed a multimodal-based FCN for graphical objects in document analysis using linguistic and visual features and improved the segmentation performance. He et al. \cite{He761} presented a multi-task and multi-scale-based FCN for predicting masks for the table, figure, text and their related contours. Then they filtered these segmentation masks by the Conditional Random field (CRF) network to get table areas in document images. In \cite{Isaak36}, the author gave a saliency-based FCN for multi-scales reasoning with CRF to detect charts and tables in digital documents. However, detection transformers can handle objects of different sizes and aspect ratios more effectively than semantic segmentation approaches, as they do not rely on fixed-size segmentation masks. This makes detection transformers more suitable for detecting objects with irregular shapes or objects that are densely packed together.\\

\noindent\textbf {Bottom-up Approaches.}~The bottom-up approaches analyze the text documents as graphs where objects (e.g., text, table) are considered nodes, and edges show the relationship between these objects to solve the table detection as a graph labelling problem. In \cite{Li64}, the author applied popular layout analysis approaches to form clusters and classify page objects (table, figure, text, formula) using CNN-based CRF networks. The overlapping regions of the same cluster and class merged to detect page objects. In \cite{Riba65, Martin66}, the authors represented text areas (text lines, words) as nodes and page layouts as graphs. They then used graph neural networks to classify edges and nodes. Finally, they extracted the table class from subgraphs where detected tables are present in the nodes. Li et al. \cite{Li49} presented a document analysis problem as a sequence-labelling problem. They classified each word in the sequence into predefined object categories, including tables, using pre-trained language models. All these approaches also need correct text/word bounding boxes as an extra input. However, detection transformers can be more efficient than bottom-up approaches for graphical page object detection, as they can process the entire image in a single forward pass through the neural network, without the need for additional post-processing steps.\\

\noindent\textbf {Object Detection-based Approaches.}~In \cite{Yi77, Oliveira565}, the authors used R-CNN \cite{rcnn13} for detecting tables. However, the performance of these methods depends on handcrafted features and heuristic rules, as in earlier approaches. The researchers \cite{Xinyi89, Saha43, Ayan29, Agarwal52} also used more advanced single-stage \cite{Joseph15,retina34} and two-stage  \cite{fast15, faster23, mask86, Cai56} object detection approaches to detect tables, formulas and figures in the document analysis domain. Further, augmentation approaches are used to boost the performance of these networks for detecting tables. As Prasad et al. \cite{Ayan29}, Arif et al. \cite{arif48} and Gilani et al. \cite{Azka62} used image transformation methods of dilation, coloration and distance transformation to increase the training data to get more information as features from input table images. In \cite{Sidd32}, the author integrated deformable convolution with deformable (Region of Interest) RoI pooling network in Faster R-CNN \cite{faster23} to make the network more efficient for the geometric transformation such as translation, rotation, reflection and dilation. Sun et al. \cite{FasterTD} enhance the localization performance by using Faster R-CNN \cite{faster23} for corner detection and adjusting the table bounding boxes by corners using a post-processing network. Anyhow, these corner boxes are handcrafted, and their size has no clear meaning that increases the false detection rate of corner boxes \cite{FasterTD}. Agarwal et al. \cite{Agarwal52} used a composite backbone network along with deformable-convolution filters in Cascade Mask R-CNN \cite{Cai56} to boost detection results in the table analysis domain. Even though these methods improve the results on many benchmark datasets \cite{icdar19, PubLayNet3, iiit13k, icdar13, icdar17, tablebank8, Paliw9}, they use large memory and have high computational complexity. Compared to CNN, transformer networks make predictions without needing non-maximum suppression, making the network more simple and efficient. Recently, Smock et al. \cite {pubtables5} used a simple detection transformer (DETR) framework for table detection and structure recognition tasks on the PubTables dataset and observed excellent results.\\  
\subsection{Object Detection with Transformers}
Previously, people used classical detectors \cite{ssd23, retinaNet68, yolox34, faster23, fast15, mask86, shehzadi1} in the object detection domain and transformer-based networks in the sequence prediction domain \cite{nlp12}. Recently, Carion et al. \cite{att75} proposed a transformer-based architecture for object detection. This Detection transformer (DETR) \cite {detr34} uses image features as learning queries and predicts the bounding boxes using bipartite graph matching. This architecture removes the need for hand-designed anchors \cite{anchors45} and non-maximum suppression (NMS) to provide a simpler and optimal object detection framework \cite{shehzadi2}. Compared with anchor-based object detectors \cite{rcnn13, fast15, faster23, mask86, sppNet57, ssd23, yolox34, retinaNet68}, transformers consider object detection a prediction problem and extract features from an image using learning queries that eliminate the need for non-maximum suppression. However, transformer-based detector like DETR has slower convergence during training than previous detectors like Faster R-CNN \cite{faster23}. For example, DETR  needs 500 epochs, while Faster R-CNN needs 12 epochs to get the same performance on the coco detection dataset.\\
Several works \cite { CondDE, dab89, Reth78, anchorDE, Deformable54, DynamicDE} have tried to find the main reason for the slow convergence of transformer-based detection networks. Some of them tried to improve the encoder-decoder architecture. Sun et al. \cite{Reth78} proposed only encoder-based DETR by considering the low performance of the attention module in the decoder network as the main reason for slow training convergence. Dai et al. \cite{DynamicDE} proposed the regions-of-interest-based dynamic decoder that uses Region of Interest (ROI). Recent works \cite{dab89, CondDE, anchorDE, Deformable54} focus on spatial positions than multiple positions for DETR queries to extract features from the image. Conditional DETR \cite{CondDE} breaks down every query into positional and content segments to have clear similarities with the spatial region in the image. Deformable-DETR \cite{Deformable54} apply deformable attention mechanism in the decoder module to improve training convergence. Efficient-DETR \cite{efficientDE} proposes a dense prediction network to take top-K object queries. But all this work uses only 2D points as anchor regions and ignores object scaling. However, DAB-DETR \cite{dab89} takes 4D coordinates as learnable queries and continuously improves them in every layer. Recently, DN-DETR \cite{dn42} indicated that the bipartite matching approach used in Hungarian loss is one of the reasons for transformer slow training convergence and presented a denoising training-based method to overcome the slow training convergence issue. Following DAB-DETR and DN-DETR, DINO \cite{dino23} proposed the Contrastive DeNoising(CDN) module that adds extra Denoising (DN) loss. All these modified versions of DETR make the transformer network faster and boost performance. We combine all these modifications for graphical object detection task and observe a remarkable performance boost.\\ 
\section{Method}
\label{sec:method}
\begin{figure*}[h]
\centering
\includegraphics[width=0.9\textwidth]{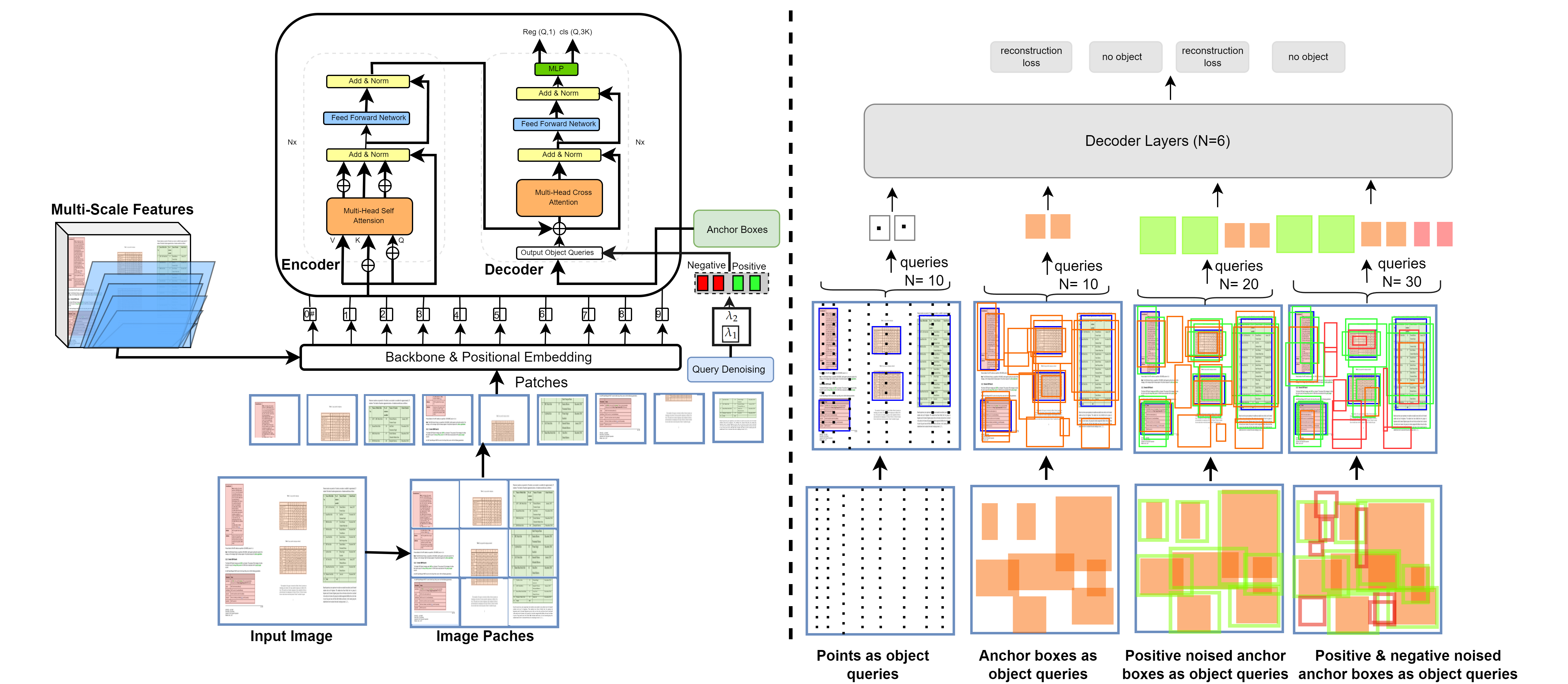}
\caption{Presented Transformer-based graphical object detection framework. We divide the input image into small equal-size patches, add position embeddings, and embed the resulting patches along with input multi-scale features to the transformer encoder. We use different forms of object queries such as points, anchor boxes, the addition of positive and negative noises to the anchor boxes in the decoder and observe network performance. Here, blue rectangles represent ground truth (GT), black dots represent points as object queries, brown rectangles denote anchor boxes as object queries, green rectangles indicate positive noised anchor boxes for foreground class, and red rectangles show negative noised anchor boxes for the background class. These object queries are taken as decoder input to provide final graphical object labels and locations.}\label{fig:encoder-decoder}
\end{figure*}

This section first describes an overview of the transformer-based detector's main modules and then employs different mechanisms to improve the object queries. The whole network is shown in Figure~\ref{fig:encoder-decoder}.
\subsection{Revisiting DETR}
\noindent\textbf {Positional Encoding and Backbone Network}
The transformer encoder takes a 1D sequence as input token embeddings. The backbone network ResNet-50 \cite{resnet45} extracts the features from the input, reduces the channel dimension, and converts the spatial dimension into one dimension as the transformer network takes input as one vector. We transform the input image $I\in R^{(H_i\times W_i\times C_i)}$ into equal size 2D patches $I_p\in R^{(N_i\times (P_i^2.C_i)}$. Where $ (H_i \times W_i)$ is the original image spatial resolution, $C_i$ is the channels/bands number, $(P_i.P_i)$ is the resolution of each image patch, $N_i= \frac{H_i.W_i}{(P_i.P_i)}$ is the total number of input patches, also considered as the transformer's input sequence length. The standard transformer model also considers any spatial relationships between the input features. So the encoder network takes two inputs: first is the feature vector of the input image as small image patches, and second is the positional encoding of the input feature vector. The resulting sequence $Z_0$ is fed as input to the transformer encoder as follows: 
\setlength{\abovedisplayskip}{3pt}
\setlength{\belowdisplayskip}{3pt}
\begin{align}
Z_0 = I_p + M_{PE} \label{eq3} 
\end{align}
Here, $I_p \in R^{(N_i \times (P_i^2.C_i)}$ is the input image patches and $M_{PE} \in R^{(N_i+1) \times D}$ represents the positional embeddings. The positional encoding contains information of the absolute or relative position of the patches to make use of the spatial information. \\
\noindent\textbf {Encoder-Decoder Network} The extracted features from the backbone network as one single vector and their position within the input vector are fed to the encoder network. Here, the self-attention layer provides key, query and value matrices which feed to the multi-head attention to find the attention probabilities of the input vector. The DETR decoder takes N number of object queries in parallel with the encoder output. During training, the model is trained to generate the same set N of predicted bounding boxes regardless of the order of the object queries. DETR searches for a permutation of N elements $\sigma \in N$ that results in the lowest cost for the matching. The cost is defined as the sum of the negative log-likelihood of the predicted class probabilities and the box loss for each matched object as follows:
\begin{equation} \label{sigma}
 \hat{\sigma} =\arg\min_{\sigma \in N} \sum_{k}^{N} \mathcal{L}_{match} (y_k,\hat{y}_{\sigma(k)}),
\end{equation}
Here, $y_k$ is a set of ground truth objects, $\hat{y}_{\sigma(k)}$ is the predicted objects. By training the model using a permutation-invariant loss, DETR is able to learn to detect objects in an image regardless of their order, which makes it more robust to variations in the input data. The term $\mathcal{L}_{match} (y_k,\hat{y}{\sigma(k)})$ is the one-to-one matching cost for direct prediction without duplicates between predicted objects and ground truth as shown in the following equation: 
\begin{equation}\label{match}
\mathcal{L}_{match} (y_k,\hat{y}_{\sigma(k)})=-\mathds{1}_{\{c_k\neq \phi\}}\hat{p}_{\sigma(k)}(c_k)+\mathds{1}_{\{c_k\neq \phi\}}\mathcal{L}_{bbox}(b_k,\hat{b}_{\sigma(k)})
\end{equation}
Where $\hat{p}_{{\sigma}(k)}$ and $c_k$ are the predicted class labels and target labels, respectively, $b_k$ and $\hat{b}_{\hat{\sigma}}(k)$ are ground truth and predicted bounding boxes, respectively.

To match the object queries N with the ground-truth objects in the image, DETR uses the Hungarian algorithm to find a bipartite matching between the set of object queries and the set of ground-truth objects. To do this, the set of ground-truth objects is padded with empty objects $(\phi)$ to make it the same size as the set of object queries. The next step is to compute the Hungarian loss $\mathcal{L}_{H}$ in Equation~(\ref{hung}) by determining the optimal matching between ground truth (GT) and detected boxes regarding bounding box location and class. It is defined as a linear combination of two terms: a negative log-likelihood term for the class predictions, and a box loss term for the predicted bounding boxes as follows.
\begin{equation}\label{hung}
 \mathcal{L}_{H}(y, \hat y)= \sum_{i=1}^{N}[-log\hat{p}_{\hat{\sigma}(k)}(c_k)+\mathds{1}_{\{c_k\neq \phi\}}\mathcal{L}_{bbox}(b_k,\hat{b}_{\hat{\sigma}}(k))]
\end{equation}
Where $\hat{\sigma}$ is the optimal-assignment factor from Equation~(\ref{sigma}). The negative log-likelihood term measures the difference between the predicted class probabilities and the ground-truth class labels. The box loss term measures the difference between the predicted bounding boxes and the ground-truth bounding boxes. This network doesn't need NMS to remove redundant predictions as it uses Hungarian loss that learns to make non-redundant predictions. However, the DETR network has several challenges, such as optimizing the network because of its slow training convergence and performance drops for small objects.
\subsection{Proposed Advancements}
\noindent\textbf {Object queries.}~~
Object queries can be viewed as a replacement for anchor boxes in object detection models, such as the popular Faster R-CNN \cite{faster23}. During training, the model learns a fixed number of object queries that represent specific object classes. In inference, detection transformer directly outputs a set of object detection using a fixed number of object queries, eliminating the need for anchor boxes and non-maximum suppression. The use of object queries enables direct, end-to-end optimization for object detection and improves interpretability of the model's outputs. We modify the object queries to observe the effect of quantity and quality of object queries on detection transformer performance for the graphical object detection task.\\
\noindent\textbf {Object queries as points.}~~Graphical object detection can be challenging due to the presence of multiple objects at the same location. For example, text, tables, and other graphical elements can overlap or appear in close proximity to each other. This issue can be resolved by using points as objects queries. We used two type of points as grid and learned points. Grid points are fixed points placed at regular intervals in the image, while learned points are initialized randomly and updated during training to better match the objects in the image. During training, the model learns a fixed number of points, which are used to define regions in the image where objects are likely to be present. The model then predicts the presence and location of objects near each point. This allows the model to predict multiple objects at one position, which is particularly useful for graphical object detection.
 
\noindent\textbf {Object queries as anchor boxes.}~~In traditional object detection tasks \cite{faster23, fast5}, anchor boxes are used as references during training and prediction to generate the final bounding box predictions. For graphical object detection such as tables, the object shapes and sizes are much more regular and predictable, since tables generally consist of rows and columns of consistent dimensions. The dynamic anchor boxes are better suited to the regular and predictable nature of table structures, and result in more accurate and reliable table boundary predictions.\\
In this case, decoder network takes positional queries as anchor boxes and needs keys $k_i$, queries $q_i$ and values $v_i$ having a cross-attention module to find features probing. Given a bounding box coordinates $a_i = (a^\prime_i, y^\prime_i , w^\prime_i, h^\prime_i )$ with content query $C_q$ to detect the contents of a table cell in a document image, its  positional query $P_i$ is formed as follows:  
\begin{equation}
 P_i= MLP (PE(a_i)) 
\end{equation}
\begin{equation}
\label{eqn:pe}
PE (a_i)=Conc(PE(x^\prime_i),PE(y^\prime_i),PE(w^\prime_i),PE(h^\prime_i)) 
\end{equation}
Here, $PE (a_i)$ is the positional encoding, it is the overall positional encoding of the bounding box coordinates by finding the positional encoding of its component and then concatenating them by the $Conc$ function. The positional query $P_i$ effectively captures the complex visual patterns present in document images, such as the varying thicknesses and styles of table borders, and the presence of different types of content within table cells.\\
\begin{figure*}[h]
\centering
\includegraphics[width=1\textwidth]{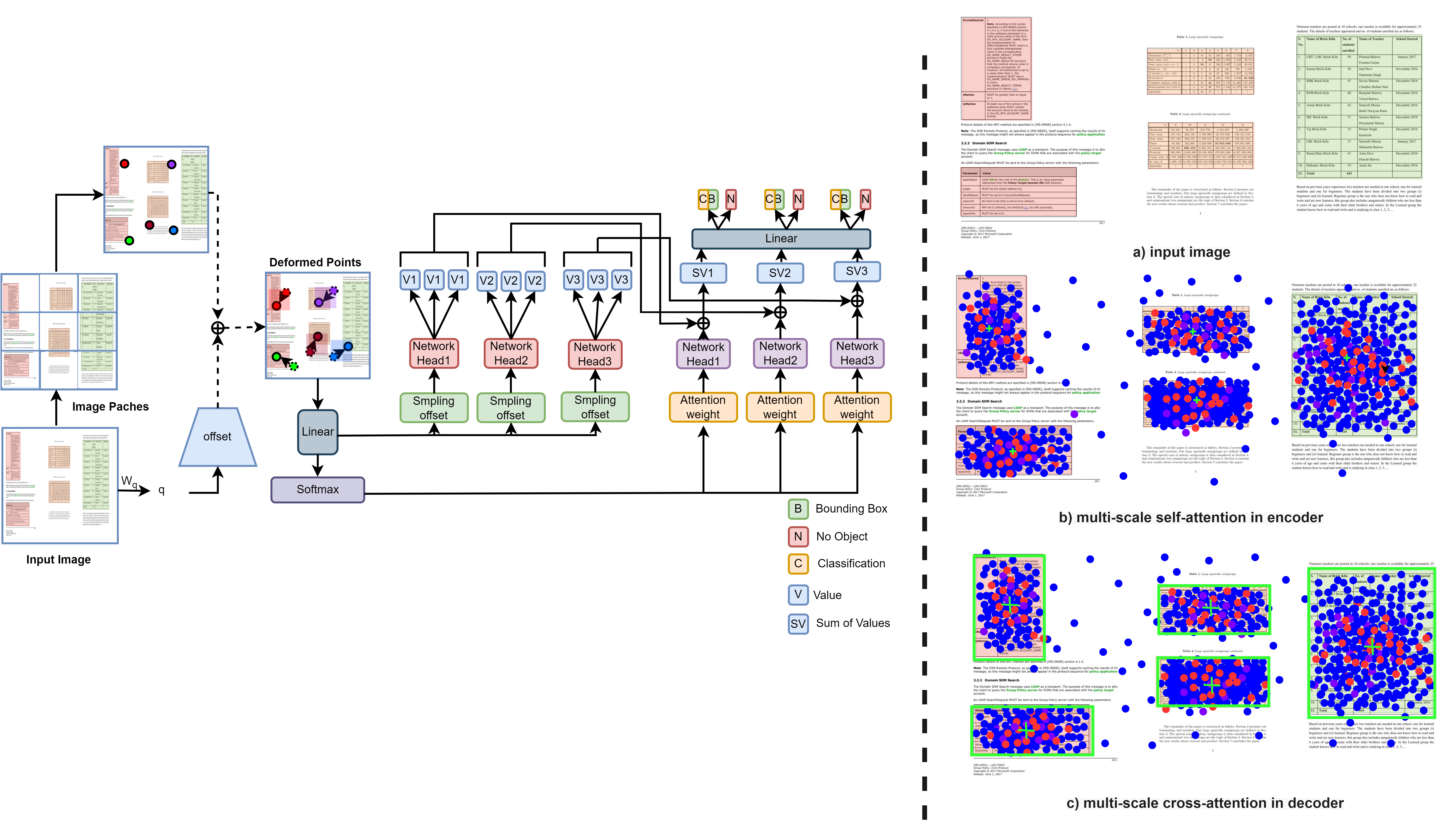}
\caption{Illustrating design of attention module. The attention network only takes a finite number of samples near the reference point, irrespective of feature map size. Here, Sampling is performed at multiple scales using a deformation field computed by a deformable convolutional neural network. For clear visualization, we show sampling points with attention weights as a circle where the circle color represents its attention weight: blue indicates low intensity while red indicates high intensity. The rectangle represents the predicted bounding box in the decoder. }\label{fig:attention}
\end{figure*} 
In the self-attention module, the queries, keys, and values all have the same content information, while key and query has extra positional information $PE (a_i)$. In cross attention, the queries are derived from the dynamic anchor boxes $a_i$, while the keys and values are derived from the feature map.
The multi-head self-attention consider deformable attention to allow the model to attend to different parts of the image at different resolutions and scales as shown in Figure~\ref{fig:attention}. Here, attention network only takes a finite number of samples near the reference point, irrespective of feature map size. Considering only a small number of keys for each query converges the network faster. \\
\noindent\textbf {Object queries as noised anchor boxes.}~
 Anchor boxes can be noisy and imprecise, leading to inaccurate detections. Adding positive noise to better fit for foreground objects and negative noise to better fit for background objects modifies the anchor boxes used in the detection process to better fit the rectangular/square shape of graphical objects. 
For this, positive noise is added to the anchor boxes to expand their size and account for small variations in object size and shape. Negative noise, on the other hand, is used for detecting background objects as shown in Figure~\ref{fig:encoder-decoder}. By combining positive and negative noise, we can more accurately capture the range of object sizes and aspect ratios present in the data while filtering out noise and irrelevant regions. $\lambda_1$ controls the amount of positive noise added to the anchor boxes, while $\lambda_2$ controls the amount of negative noise. These values are learned during training and are used to adjust the size of the anchor boxes, allowing them to better match the objects in the image. The initial anchor boxes at the decoder input is represented as $a_i$, where $a_i=(x^\prime_i,y^\prime_i,w^\prime_i,h^\prime_i)$, while we have N number of GT boxes represented as $b_i$ where $b_i=(x_i,y_i,w_i,h_i)$. We remove those anchors that are farther away from the ground truth anchor as follows: 
\begin{equation}
 AMD(k) = \frac{1}{k} \Sigma \{Max_K (\{\parallel b_0-a_0\parallel_1,...,\parallel b_{N-1}-a_{N-1}\parallel_1\},k )\}
\end{equation}
 Where $(b_i-a_i)$ is the distance between bounding box b and anchor and $Max_K$ is the module that selects the top K anchors. Usually, the value of $\lambda_2$ is small to improve model performance as hard negative examples are closer ground truth boxes. If an image has N number of objects to be detected, we will have a total of 3N queries generating 2N positive and negative anchors for N ground truths. \\
\subsection{Pre-processing Methods}
 We use some transformation techniques that help model learning more accurately during training. Document images contain blank spaces and content or text regions. As the network detects graphical objects in document data, we apply different transformation techniques to form the text or table regions thicker and reduce the blank space regions. For that, we use two types of approaches as smudge transform and dilation transform.\\

\begin{figure}
\centering
\includegraphics[width=.9\textwidth]{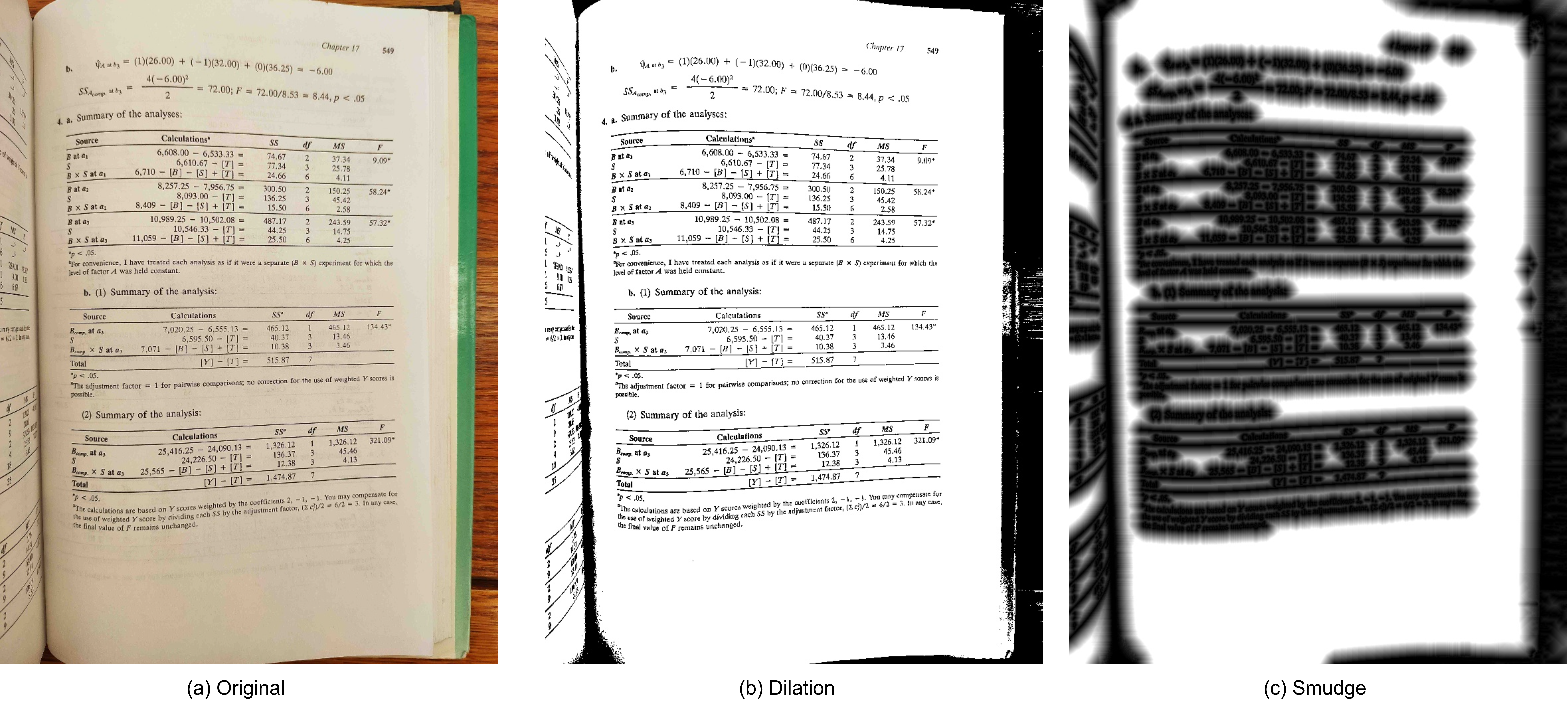}
\caption{Pre-processing techniques to
form the text or table regions thicker and reduce the blank space regions. Here, (a) shows the original data sample, (b) contains a dilation transformation-based augmentation sample, and (c) provides a smudge transformation-based augmentation sample.}\label{fig:trans}
\end{figure}

\noindent\textbf{Dilation Transform.}~
In this transformation, black pixel regions are thicker by transforming the original table image. First, we convert the original image into a binary image and then apply the dilation transform by a kernel filter (2x2 size) for one iteration. We select this kernel size because it gives the best results. The original image is represented as a), and the image after dilation is represented as b) in Figure~\ref{fig:trans}. \\

\noindent\textbf{Smudge Transform.}~
In this transformation, black pixel regions spread around the end of the black regions. First, we convert the original image into a binary image and then apply the smudge transformation by various distance transforms. Gilani et al. \cite{Azka62} explained the original algorithm that applies Linear Distance Transform, Max Distance Transform and Euclidean Distance Transform to the input image. Furthermore, We also normalize and tune the parameters to boost the results. The original image is represented as a), and the image after the smudge transformation is represented as c) in Figure~\ref{fig:trans}. 

%%%%%%%%%%%%%%%%%%%%%%%%%%%%%%%%%%%%%%%%%%
\section{Experiments }
\label{sec:experiments_results}
\subsection{Dataset and Evaluation Setup}
The performance of the transformer-based table detection model is evaluated using precision, recall, F-Score and mean average precision (mAP) in the context of MS COCO \cite{coco14} evaluation on the four most extensive graphical object detection datasets: TableBank \cite{tablebank8}, PubLayNet \cite{PubLayNet3}, PubTables \cite{pubtables5} and NTable \cite{NTables} dataset. Please refer to the supplementary material for a detailed explanation of the dataset and evaluation setup.

\subsection{Results}   
\subsubsection{TableBank}
\label{sec:TableBank-results}
\begin{figure}[h!]
\centering
\includegraphics[width=1\textwidth]{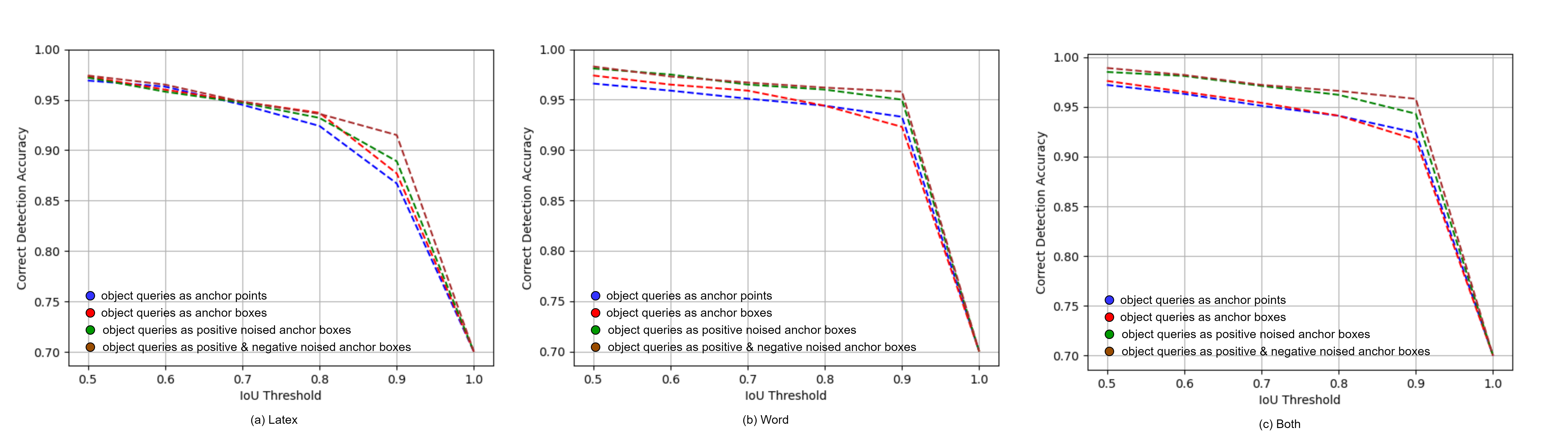}
\caption{Performance analysis of DETR and modifications in object queries in terms of AP over the IoU thresholds range from 0.5 to 1.0 on word, latex and both splits of the TableBank dataset. Here, the blue color shows results with simple DETR, the red highlights results with anchor boxes as object queries, the green represents results with positive noised anchor boxes, and the brown denotes results with positive and negative noised anchor boxes.}\label{fig:TableBank-iou}
\end{figure}

\begin{figure}[h!]
\centering
\includegraphics[width=1\textwidth]{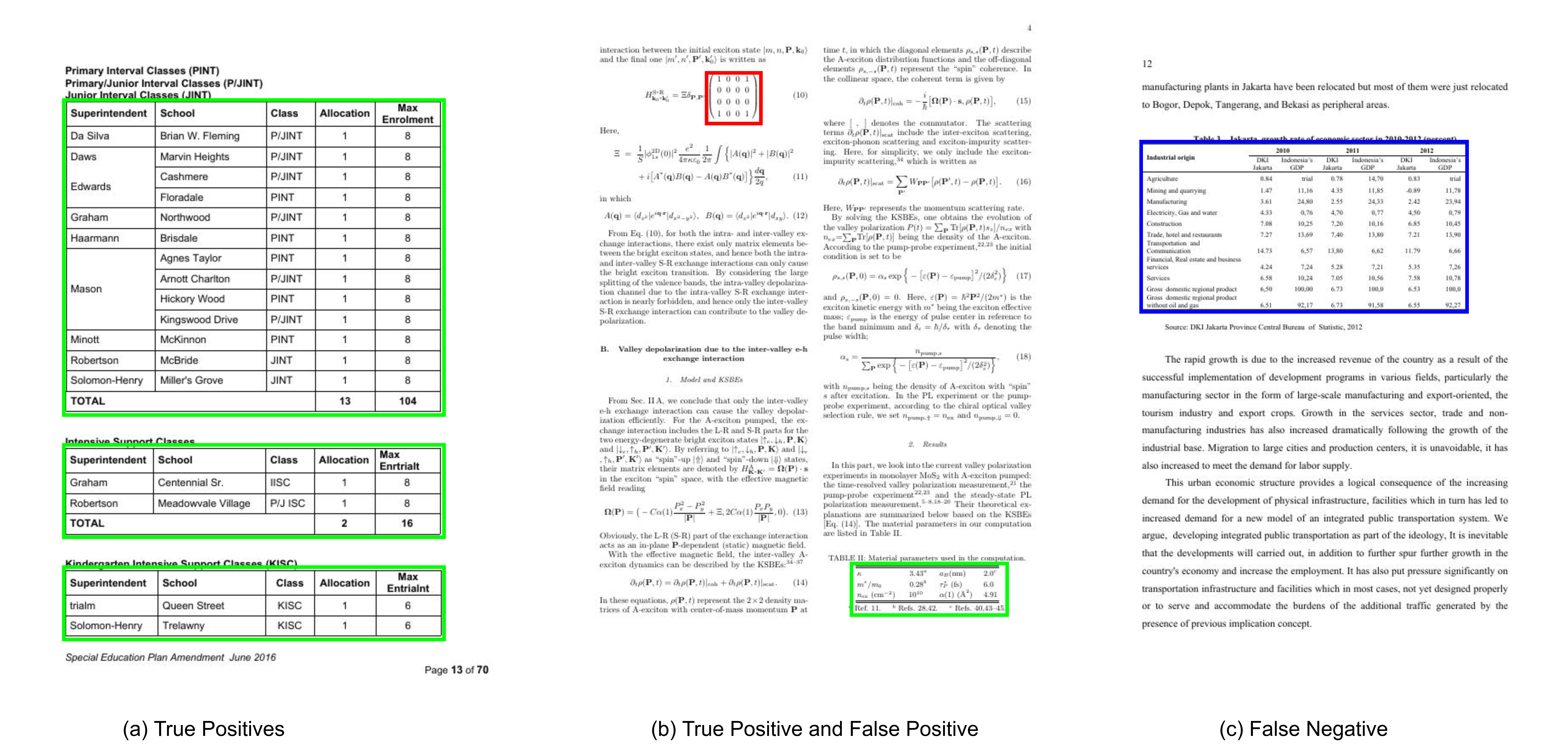}
\caption{Transformer-based detector with improved object queries (by adding positive and negative noised anchors) results on TableBank$_{both}$ dataset. Green color denotes true positives, blue exhibits false negatives and red exhibits false positives. Here, (a) exhibits true positive detection samples, (b) contains true positive and false positive detection samples, and (c) provides false negative detection.}\label{fig:TableBank-result}
\end{figure}

We validate the performance of detection transformer by modifying the object queries on the raw (without pre-processing), dilation and smudge transformation of the TableBank dataset in Table~\ref{tab:TableBank1}. Here, the term "both" represents combination of latex+word split.  
\begin{table*}
\tiny
\begin{center}
\caption{Comparison between transformer-based detector results on raw (without pre-processing), dilation and smudge transformation of the TableBank dataset. Here, term $Q_b$ represents object queries as anchor boxes, $Q_p$ denotes object queries with positive noise and $Q_n$ indicates object queries with negative noise. The IoU thresholds are set to 0.5 and 0.75 for average precision calculation and also calculate average recall for large objects. AR represents Average Recall for a large area. The best results are highlighted.}\label{tab:TableBank1}
\renewcommand{\arraystretch}{1} % Default value: 1
\begin{tabular*}{\textwidth}{@{\extracolsep{\fill}}lccccc@{\extracolsep{\fill}}}
\toprule
\textbf{Methods} &
\textbf{Preprocessing} &
\textbf{mAP} & 
\textbf{AP\textsuperscript{50}} &
\textbf{AP\textsuperscript{75}}  & \textbf{AR} \\
\midrule
\multirow{2}{*}{DETR} & raw(latex) & 86.6& 96.9 & 93.7& 92.6\\
& Dilation+Smudge & 87.2$\pm$0.21 & 97.5 & 94.2 & 93.5\\
\midrule

\multirow{2}{*}{DETR + $Q_b$} & raw (latex)  & 87.7 & 97.3 & 94.8  & 93.5\\
& Dilation+Smudge  & 88.2$\pm$0.53 & 98.0 & 95.4 & 94.3\\
\midrule

\multirow{2}{*}{DETR + $Q_b$ + $Q_p$} & raw (latex)  & 88.9 & 97.2 & 94.7 & 94.6 \\
& Dilation+Smudge  & 89.4$\pm$0.61 & 97.6 & 94.9 & 95.4\\
\midrule

\multirow{2}{*}{\textbf{DETR + $Q_b$ + $Q_p$ + $Q_n$}} & \textbf{raw (latex)} & \textbf{91.5} & \textbf{97.4} & \textbf{94.8} & \textbf{97.8}\\
& \textbf{Dilation+Smudge} & \textbf{92.6$\pm$1.23} & \textbf{98.2} & \textbf{95.0} & \textbf{98.4}\\
\midrule

\multirow{2}{*}{DETR} & raw(word)  & 93.4& 96.6 & 95.1 & 96.7\\
& Dilation+Smudge  & 93.9$\pm$0.31 & 97.3 & 95.9  & 97.2\\
\midrule

\multirow{2}{*}{DETR + $Q_b$} & raw (word) & 92.3 & 97.4 & 95.9  & 95.0\\
& Dilation+Smudge & 93.1$\pm$0.40 & 97.9 & 96.4  & 95.8\\
\midrule

\multirow{2}{*}{DETR + $Q_b$ + $Q_p$} & raw (word) & 95.0 & 98.1 & 96.5 & 96.6 \\
& Dilation+Smudge & 95.5$\pm$1.5 & 98.7 & 96.9 & 96.9\\
\midrule

\multirow{2}{*}{\textbf{DETR + $Q_b$ + $Q_p$ + $Q_n$}} & \textbf{raw (word)} & \textbf{96.3} & \textbf{98.3} & \textbf{96.5} & \textbf{99.1}\\
& \textbf{Dilation+Smudge} & \textbf{96.8$\pm$1.24} & \textbf{98.8}& \textbf{96.9}& \textbf{99.7}\\
\midrule

\multirow{2}{*}{DETR}& raw (both)& 92.5 & 97.2 & 95.1 &  96.8\\
& Dilation+Smudge & 93.4$\pm$1.1 & 97.5 & 95.8 & 97.3\\
\midrule

\multirow{2}{*}{DETR + $Q_b$} & raw (both)& 91.7 & 97.6 & 95.4 & 96.4\\
& Dilation+Smudge & 95.1$\pm$1.23 & 97.8 & 96.9  & 97.4\\
\midrule

\multirow{2}{*}{DETR + $Q_b$ + $Q_p$} & raw (both) & 94.3 & 98.5 & 97.1 & 97.7 \\
& Dilation+Smudge & 94.9$\pm$0.5 & 98.8 & 97.9 & 98.8\\
\midrule

\multirow{2}{*}{\textbf{DETR + $Q_b$ + $Q_p$ + $Q_n$}} & \textbf{raw (both)} & \textbf{95.8} & \textbf{98.9} & \textbf{97.2} & \textbf{98.8}\\
& \textbf{Dilation+Smudge} & \textbf{96.9$\pm$0.41} & \textbf{99.4} & \textbf{97.6} & \textbf{99.1}\\

\bottomrule
\end{tabular*}
\end{center}
\end{table*} 
For the data group of TableBank$_{both}$ with dilation and smudge transformation, we achieve an mAP of 93.4$\%$ for grid points used as object queries, 95.1$\%$ for anchor boxes as object queries, 94.9$\%$ for positive noised anchor boxes as object queries and 96.9$\%$ for positive and negative noised anchor boxes as object queries. \\
Figure~\ref{fig:TableBank-iou} exhibits the performance analysis of detection transformers using different types of object queries as input using Average Precision (AP) with IoU threshold values ranging from 0.5 to 1 on all splits TableBank dataset. We can observe that detection transformer that uses points as object queries shows the lowest performance, while noised anchor boxes as object queries has the highest result on all threshold values. The qualitative analysis of table detection for the TableBank$_{both}$ dataset is illustrated in Figure~\ref{fig:TableBank-result}. Analysis of incorrect results reveals that the network fails to localize accurate tabular areas or gives false positives.\\

\noindent\textbf{Comparison with State-of-the-art Methods}
We also compare the results of the detection transformer using different type of object queries with earlier CNN-based approaches on the TableBank dataset. Table~\ref{tab:TableBank2} shows that transformer with noised anchor boxes has outperformed the previous state-of-the-art methods.  \par

\begin{table*}
\tiny
\begin{center}
\caption{Comparison between the transformer-based detectors and previous state-of-the-art results on the TableBank dataset \colorbox{red!25}{without pre-processing} (raw data). Here, term $Q_b$ represents object queries as anchor boxes, $Q_p$ denotes object queries with positive noise and $Q_n$ indicates object queries with negative noise. The IoU threshold value is set to 0.5. The best results are highlighted.}\label{tab:TableBank2}%
\begin{tabular*}{\textwidth}{@{\extracolsep{\fill}}cccc@{\extracolsep{\fill}}}
\toprule
\textbf{Model} &
\textbf{Split} &
\textbf{AP\textsuperscript{50}} &
\textbf{AR} \\ 
\midrule
 CascadeTabNet \cite{Ayan29} &
 TableBank$_{latex}$ &
  95.9 &
  97.2 \\ \midrule
  HybridTabNet \cite{HybridTabNet}  &
  TableBank$_{latex}$&
  97.7 &
  98.3  \\ \midrule
  CasTabDetectoRS \cite{CasTab45}  &
  TableBank$_{latex}$ &
  98.3 &
  98.4  \\ \midrule
  \textbf{DETR + $Q_b$ + $Q_p$ + $Q_n$} &
  \textbf{TableBank$_{latex}$}&
  \textbf{97.4} &
  \textbf{97.8}  \\ \midrule
 CascadeTabNet \cite{Ayan29}  &
 TableBank$_{word}$&
  94.3  &
  95.5 \\ \midrule
  HybridTabNet \cite{HybridTabNet} &
  TableBank$_{word}$&
  95.5 &
  98.5   \\ \midrule
  CasTabDetectoRS \cite{CasTab45}  &
  TableBank$_{word}$ &
  96.7 &
  98.5  \\ \midrule
 \textbf{DETR + $Q_b$ + $Q_p$ + $Q_n$} &
 \textbf{TableBank$_{word}$}&
 \textbf{98.3} &
 \textbf{99.1}\\
 \midrule
 CascadeTabNet \cite{Ayan29} &
 TableBank$_{both}$ &
  94.4  &
  95.7 \\ \midrule
  HybridTabNet \cite{HybridTabNet}   &
 TableBank$_{both}$&
  96.3 &
  98.6  \\ \midrule
  CasTabDetectoRS \cite{CasTab45} &
   TableBank$_{both}$ &
   97.4&
  98.2   \\ \midrule
 \textbf{DETR + $Q_b$ + $Q_p$ + $Q_n$} &
 \textbf{TableBank$_{both}$}&
 \textbf{0.989} &
 \textbf{0.988}  \\ 
\bottomrule
\end{tabular*}
\end{center}
\end{table*} 

\subsubsection{PubLayNet}
\label{sec:PubLayNet-result}
\begin{figure}[h!]
\centering
\includegraphics[width=1\textwidth]{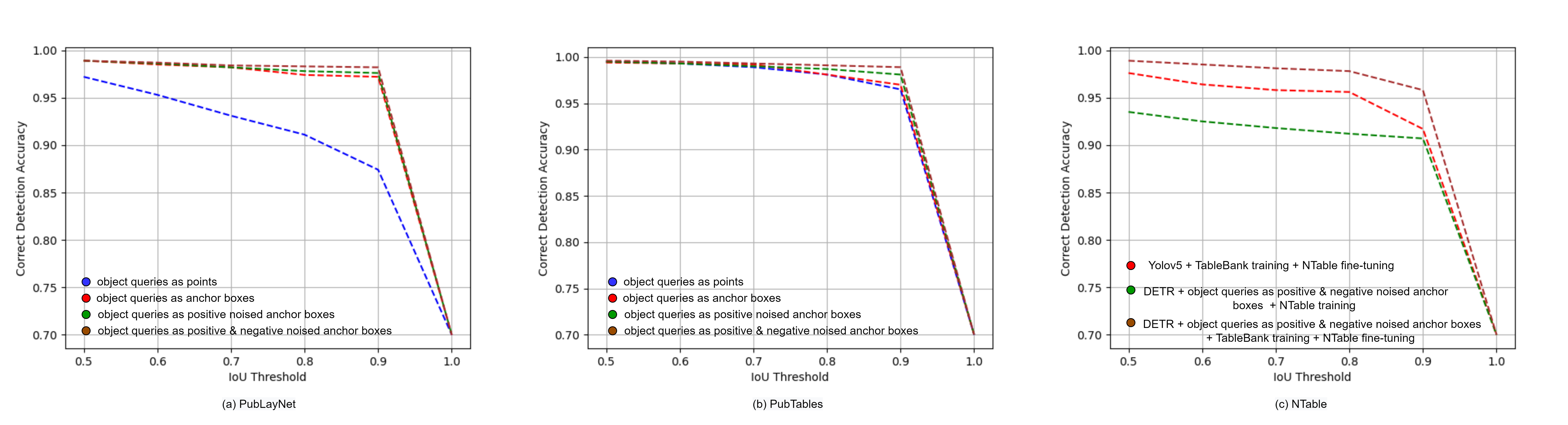}
\caption{Performance analysis of DETR with modifications in object queries in terms of AP over the IoU threshold values range from 0.5 to 1.0 on the a) PubLayNet, b) PubTables and c) NTable datasets. In part a) and b) of this figures, the blue color shows results with simple DETR, the red highlights results with anchor boxes as object queries, the green represents results with positive noised anchor boxes, and the brown denotes results with positive and negative noised anchor boxes. In part c), we compare results with previous YOLOv5 framework on NTable dataset for camera-based table images.}\label{fig:PubLayNet-pubtables-iou}
\end{figure}

We validate the performance of detection transformer by modifying the object queries on the raw, dilation and smudge transformation of the PubLayNet dataset in Table~\ref{tab:PubLayNet1}. we consider all PubLayNet classes (table, title, figure, list and text). We get the best results having mAP of 95.7$\%$  with noised anchor boxes taken as object queries. Anchor boxes are typically used in object detection models to help the model localize and classify objects in an image. However, in the context of graphical page object detection, anchor boxes used are too small or too large, they may not capture the full extent of an object on the page, leading to incorrect predictions. Noisy anchor boxes can help to capture more diverse object shapes and sizes, especially when the objects in the images have a wide variety of aspect ratios and sizes. By introducing random noise into the anchor boxes, the model is forced to learn more robust features and adapt to a wider range of object sizes and shapes, which can improve its overall performance.  \\
\begin{table*}[h!]
\tiny
\begin{center}
\caption{Comparison between DETR and its submodules on raw
(without pre-processing), dilation and smudge transformation of the PubLayNet dataset. Here, term $Q_b$ represents object queries as anchor boxes, $Q_p$ denotes object queries with positive noise and $Q_n$ indicates object queries with negative noise. The IoU thresholds are set to 0.5 and 0.75. AR represents average precision for a large area. The best results are highlighted.}\label{tab:PubLayNet1}
\begin{tabular*}{\textwidth}{@{\extracolsep{\fill}}lccccc@{\extracolsep{\fill}}}
\toprule
\textbf{Methods} & \textbf{Preprocessing} &
\textbf{mAP} & 
\textbf{AP\textsuperscript{50}} &
\textbf{AP\textsuperscript{75}} & \textbf{AR} \\
\midrule
\multirow{2}{*}{DETR} & raw & 93.6 & 95.3 & 94.1 & 94.5\\
%\hhline{~-----}
& Dilation+Smudge & 93.9$\pm$0.12 & 95.6 & 94.4  & 94.7\\
\midrule

\multirow{2}{*}{DETR + $Q_b$} & raw & 94.2 & 95.8 & 94.6 & 94.9\\
%\hhline{~-----}
& Dilation+Smudge & 94.7$\pm$0.31 & 96.0 & 95.0  & 95.1\\
\midrule

\multirow{2}{*}{DETR + $Q_b$ + $Q_p$} & raw & 94.8 & 96.3 & 95.5 & 95.3 \\
%\hhline{~-----}
& Dilation+Smudge & 95.0$\pm$0.10 & 96.5 & 95.9  & 95.5\\
\midrule

\multirow{2}{*}{DETR + $Q_b$ + $Q_p$ + $Q_n$} & \textbf{raw} & \textbf{95.2} & \textbf{96.9} & \textbf{96.2} & \textbf{95.7}\\
& \textbf{Dilation+Smudge} & \textcolor{red}{95.7$\pm$0.11} & \textcolor{red}{97.5} & \textcolor{red}{96.7} & \textcolor{red}{96.9}\\
\bottomrule
\end{tabular*}
\end{center}
\end{table*} 
The qualitative analysis for the PubLayNet dataset is illustrated in (a) part of Figure~\ref{fig:PubLayNet-pubtables-iou}. Here, the performance (AP) using noised anchor boxes as object queries is highest on all IoU threshold values represented with a brown dotted line. \\
\noindent\textbf{Comparison with state-of-the-art methods.}~
We also compare the results of transformer with modifications in object queries with earlier approaches on the PubLayNet dataset. Table~\ref{tab:PubLayNet2} shows that transformer with noised anchor boxes as object queries has outperformed the previous state-of-the-art methods for detecting graphical objects in document images.\\
\begin{table*}
\tiny
\begin{center}
\caption{Comparison between the transformer-based detectors and previous state-of-the-art results on PubLayNet validation set \colorbox{red!25}{without pre-processing} (raw data). The term $Q_b$ represents object queries as anchor boxes, $Q_p$ denotes object queries with positive noise and $Q_n$ indicates object queries with negative noise. Here, the mAP is for all these graphical objects. The best results are exhibited.}\label{tab:PubLayNet2}%
\begin{tabular*}{\textwidth}
{@{\extracolsep{\fill}}lp{1.7cm}lllllll@{\extracolsep{\fill}}}
\toprule
\textbf{Model} &
\textbf{Framework} &
\textbf{Backbone} &
\textbf{Table} &
\textbf{Text} &
\textbf{Title} &
\textbf{List} &
\textbf{Figure} &
\textbf{mAP} \\
\midrule
PubLayNet \cite{PubLayNet3} &
Mask R-CNN  &
ResNet-101 &
96.0 & 91.6 & 84.0 & 88.6 & 94.9 & 91.0 \\ \midrule

UDoc \cite{UDoc39} &
Faster R-CNN &
ResNet-50 &
97.3 & 93.9 & 88.5 & 93.7 & 96.4 & 93.9 \\ \midrule

DiT$_{B}$ \cite{Lidit78} & 
Cascade R-CNN &
Transformer &
97.6 & 94.4 & 88.9 & 94.8 & 96.9 & 94.5\\ \midrule

LayoutLMv3$_{B}$ \cite{layoutMV3} &
Cascade R-CNN &
 Transformer &
97.9 & 94.5 & 90.6 & 95.5 & 97.0 & 95.1 \\ \midrule

\textbf{Our } &
\textbf{DETR+$Q_b$+$ $ $Q_p$+$Q_n$} &
\textbf{ResNet-50} &
\textcolor{red}{98.1} &
\textcolor{red}{94.7} &
\textcolor{red}{91.8} &
\textcolor{red}{96.4} &
\textcolor{red}{97.5} &
\textcolor{red}{95.7} \\
\bottomrule
\end{tabular*}
\end{center}
\end{table*}

\subsubsection{PubTables}
\label{sec:PubTables-result}
PubTables is the largest table dataset on which we evaluate the capabilities of detection transformer and modifications in object queries. Table~\ref{tab:PubTables1} shows the results of all modules on the raw (without pre-processing), dilation and smudge transformation of the PubTables dataset. It shows that the nosied anchor boxes as object queries shows the best results.

\begin{table*}[h!]
\tiny
\begin{center}
\caption{Comparison between transformer-based detector results on  raw (without pre-processing), dilation and smudge transformation of the PubTables dataset. Here, term $Q_b$ represents object queries as anchor boxes, $Q_p$ denotes object queries with positive noise and $Q_n$ indicates object queries with negative noise. The IoU thresholds are set to 0.5 and 0.75 for average precision and also calculate average recall for large objects. The best results are highlighted. }\label{tab:PubTables1}
\begin{tabular*}{\textwidth}{@{\extracolsep{\fill}}lccccc@{\extracolsep{\fill}}}
\toprule
\textbf{Methods} & 
\textbf{Preprocessing } &
\textbf{mAP} & 
\textbf{AP\textsuperscript{50}} &
\textbf{AP\textsuperscript{75}} &
\textbf{AR} \\
\midrule
\multirow{2}{*}{DETR} & raw & 97.6& 99.5 & 98.9  & 98.5\\
& Dilation + Smudge & 97.8$\pm$0.11 & 99.8 & 99.3 & 98.9\\

\midrule
 \multirow{2}{*}{DETR + $Q_b$ }& raw & 98.0 & 99.4 & 99.2 & 99.2\\
& Dilation + Smudge & 98.6$\pm$0.21 & 99.8& 99.5 & 99.5\\
\midrule

\multirow{2}{*}{DETR + $Q_b$ + $Q_p$} & raw & 98.1 & 99.5 & 99.0 & 99.2\\
& Dilation + Smudge & 98.6$\pm$0.02 & 99.8 & 99.5 & 99.5\\
\midrule
\multirow{2}{*}{\textbf{DETR + $Q_b$ + $Q_p$ + $Q_n$}}& \textbf{raw} & \textbf{98.9} & \textbf{99.6} & \textbf{99.3} & \textbf{99.4}\\
& \textbf{Dilation + Smudge} & \textcolor{red}{99.3$\pm$0.42} & \textcolor{red}{99.9} & \textcolor{red}{99.8} & \textcolor{red}{99.6}\\
\bottomrule
\end{tabular*}
\end{center}
\end{table*} 

The qualitative analysis on the PubTables dataset is illustrated in (b) part of Figure~\ref{fig:PubLayNet-pubtables-iou}. It provides the Average precision (AP) on all IoU thresholds. The improved query network, represented with a brown dotted line, gives the highest mAP on all IoU threshold values. 

\noindent\textbf{Comparison with State-of-the-art Methods}
Recently, Smock et al.\cite {pubtables5} show an 82.5$\%$ mAP, 98.5$\%$ AP at 0.5 and 92.7$\%$ AP at 0.75 IoU threshold on the PuTables dataset on the Faster R-CNN detector as shown in Table~\ref{tab:pubtable}. As the PubTables dataset is newly released, no previous work on this dataset is available.

\begin{table*}
\tiny
\begin{center}
\caption{Comparison between the transformer-based detectors and previous state-of-the-art results on PubTables dataset \colorbox{red!25}{without pre-processing} (raw data). Here, term $Q_b$ represents object queries as anchor boxes, $Q_p$ denotes object queries with positive noise and $Q_n$ indicates object queries with negative noise. The best results are exhibited.}\label{tab:pubtable}%
\begin{tabular*}{\textwidth}
{@{\extracolsep{\fill}}cccccccc@{\extracolsep{\fill}}}
\toprule
\textbf{Model} &
\textbf{Framework} &
\textbf{mAP} &
\textbf{AP\textsubscript{50}} &
\textbf{AP\textsubscript{75}} &
\textbf{AR}\\
\midrule

Smock et al.\cite {pubtables5} & Faster R-CNN & 82.5 & 98.5 & 92.7 & 86.6 \\

\midrule
Smock et al.\cite {pubtables5} & DETR & 96.6 & 99.5 & 98.8 & 98.1 \\ 
\midrule 
Our & DETR+ $Q_b + Q_p + Q_n$ & \textcolor{red}{98.9} & \textcolor{red}{99.6} & \textcolor{red}{99.3} & \textcolor{red}{99.4}\\

\bottomrule
\end{tabular*}
\end{center}
\end{table*}

\subsubsection{NTable}
\label{sec:NTables-result}
We evaluate modifications in object queries on camera-based table detection task using NTable dataset. In Table~\ref{tab:ntable}, we have AP value of 92.5$\%$ at IoU threshold of 0.6 trained on NTable-cam and NTable-gen data split. 
\begin{table*}[htp!]
\tiny
\begin{center}
\caption{Comparison between the transformer-based detectors and previous state-of-the-art results on NTable dataset \colorbox{red!25}{without pre-processing} (raw data). Here, term $Q_b$ represents object queries as anchor boxes, $Q_p$ denotes object queries with positive noise and $Q_n$ indicates object queries with negative noise. The best results are exhibited.}\label{tab:ntable}%
\begin{tabular*}{\textwidth}
{@{\extracolsep{\fill}}cccccccc@{\extracolsep{\fill}}}
\toprule
\textbf{Model} &
\textbf{Framework} &
\textbf{Train} &
\textbf{Fine-tuning} &
\textbf{IoU} &
\textbf{AP} &
\textbf{AR} &
\textbf{F-Score} \\
\midrule

\multirow{2}{*}{NTable \cite{NTables}} & \multirow{2}{*}{YOLOv5} & \multirow{2}{*}{TableBank} & Ntable-cam +  & 0.6 & 96.4 & 99.7 & 98.0 \\
&  &  & Ntable-gen & 0.8 & 95.6 & 98.8 & 97.2\\
\midrule

\multirow{2}{*}{Our} & \multirow{2}{*}{DETR} & \multirow{2}{*}{TableBank} & Ntable-cam +  & 0.6 & 95.9 & 99.3 & 97.6 \\
&  &  & Ntable-gen & 0.8 & 94.9 & 98.4 & 96.6\\
\midrule

\multirow{2}{*}{Our} & DETR + $Q_b$+  & NTable-cam+  & \multirow{2}{*}{-} & 0.6 & \textcolor{blue}{92.5} & 98.2 & 95.3 \\
& $Q_p$ + $Q_n$ & NTable-gen & & 0.8 & 91.2 & 97.3 & 94.2 \\
\midrule

\multirow{2}{*}{Our} & DETR+$Q_b$+ & \multirow{2}{*}{TableBank} & NTable-cam +  & 0.6 & \textcolor{red}{98.5} & \textcolor{red}{99.8} & \textcolor{red}{99.1} \\
& $Q_p$ + $Q_n$ &  & NTable-gen & 0.8 & \textcolor{red}{97.8} & \textcolor{red}{99.2} & \textcolor{red}{98.5} \\

\bottomrule
\end{tabular*}
\end{center}
\end{table*}

We train the detection transformer network with noised anchors as object queries on TableBank$_{both}$ data split and then fine tune it on NTable-cam and NTable-gen dataset. It achieves AP value of 98.5$\%$ at IoU threshold of 0.6. We also compare these results with earlier approach on NTable dataset.The qualitative analysis for the NTable dataset is illustrated in (c) part
of Figure~\ref{fig:PubLayNet-pubtables-iou}. Here, traditional anchor boxes use a fixed set of scales and aspect ratios chosen based on prior knowledge or heuristics. However, these fixed anchor boxes may not be optimal for datasets with a wide range of object sizes and shapes, such as the NTable dataset. Dynamic noised anchor boxes address this limitation by adjusting the anchor box scales and aspect ratios during training based on the dataset's distribution of object sizes and shapes.

\subsection{Ablation Studies}
\label{sec:Ablation Studies}
In this section, we perform a series of ablation studies as pre-processing, quality and quantity of object queries on network performance. This ablation study is performed on PubTables dataset.\\
\noindent\textbf{Influence of pre-processing modules.}~We study the effect of transformation approaches used in our method on performance. In Table~\ref{tab:aug-result}, the best performance is achieved using the dilation transform and smudge transform together on the table data. The results are on PubTables dataset with positive and negative noised anchor boxes as object queries. These approaches make the table regions more prominent, which increases the performance. \\
\begin{table}[htp!]
\tiny
\begin{minipage}[b]{.55\textwidth}
\caption{Effectiveness of Pre-processing as smudge and dilation transformation on network performance. }\label{tab:aug-result}
\begin{tabular}{c|c|c|c|c}
\toprule
  \textbf{Dilation} &
  \textbf{Smudge} &
  \textbf{mAP} &
  \textbf{AP\textsubscript{50}} &
  \textbf{AP\textsubscript{75}}  \\ \midrule
  \color{red}\xmark &
  \color{red}\xmark &
  98.9 &
  99.6 &
  99.3 \\ 
  \greencheck&
  \color{red}\xmark&
  99.0&
  99.7 &
  99.5\\ 
  \color{red}\xmark&
  \greencheck &
  99.2 &
  99.8 &
  99.7 \\ 
 \greencheck &
 \greencheck  &
 \textbf{99.3} &
 \textbf{99.9} &
 \textbf{99.8} \\ 
\botrule
\end{tabular}
\end{minipage}\qquad
\begin{minipage}[b]{0.45\textwidth}
\caption{Effectiveness of the number of object queries on network performance. }\label{tab:num-result}
\begin{tabular}{c|c|c|c}
\toprule
\textbf{N} &
\textbf{mAP} &
\textbf{AP\textsubscript{50}} &
\textbf{AP\textsubscript{75}}  \\
\midrule
5 &
97.0 &
98.7 &
97.0 \\ 
\textbf{ 10} & 
\textbf{98.9} &
\textbf{99.6} &
\textbf{99.3}\\  
50 &
98.7&
99.5 &
99.2\\ 
100 &
98.6 &
99.4 &
99.0\\ 
\botrule
\end{tabular}
\end{minipage}
\end{table}

\noindent\textbf{Influence of learnable object queries quantity}
We study the effect of the quantity of learnable queries on detection transformer performmance. Table~\ref{tab:num-result} shows different numbers of object queries and their effect on the performance. These results are on PubTables dataset with positive and negative noised anchor boxes as object queries without pre-processing on raw data. The best performance is achieved by setting the value of N to 10, lower or higher values will cause a significant performance drop. By setting a lower value of N, the model may not provide boxes to all objects and reduce performance by classifying some objects as false negatives. Similarly, by setting a large value of N, the model will overfit and classify no object regions as false positives that cause performance drop. \\
\noindent\textbf{Influence of object queries quality.}~ We analyze the effect object queries quality on the detection transformer performance. These object queries can be either points, anchor boxes, positive noised anchor boxes or positive and negative noised anchor boxes. 
In Table~\ref{tab:box}, We can observe that noised anchor boxes as object queries increase the performance as it gives better spatial prior for the attention module. Adding positive and negative noise to anchor boxes improves models' performance by introducing random perturbations during training to make the model more robust to variations in object size, aspect ratio, and position. Positive noise improves mAP to 98.1$\%$. 
\begin{table}[h!]
\tiny
\begin{center}
\centering
\caption{Influence of modifying object queries as points, anchor boxes, positive noised anchor boxes and positive \& negative noised anchor boxes as object queries on network performance. These are the results on Pubtables dataset on raw data without pre-processing.}\label{tab:box}%
\begin{tabular}{ccccccc}
\toprule
  \textbf{Points} &
  \textbf{Boxes} &
  \textbf{Positive noise} &
  \textbf{Negative noise} &
  \textbf{mAP} &
  \textbf{AP\textsubscript{50}} &
  \textbf{AP\textsubscript{75}}  \\ 
  \midrule
   \greencheck &
  \color{red}\xmark &
  \color{red}\xmark &
  \color{red}\xmark &
97.0 &
99.5 &
98.9\\ 
   \midrule
\color{red}\xmark &
\greencheck &
\color{red}\xmark &
\color{red}\xmark &
98.0 &
99.4 &
99.2 \\ 
\midrule
\color{red}\xmark &
\greencheck &
\greencheck &
\color{red}\xmark &
98.1 &
99.5 &
99.0 \\ 
\midrule
\color{red}\xmark &
\greencheck &
\greencheck &
\greencheck &
 98.9 &
 99.6 &
 99.3 \\
\bottomrule
\end{tabular}
\end{center}
\end{table}
 By combining positive and negative noise, the model becomes more robust to object appearance and position variations and can better differentiate between foreground and background regions, improving overall accuracy and robustness. Positive and negative noised anchors improve the mAP to 98.9$\%$. This training produces more anchors by adding noise and selects the best ones closer to ground truth that improves performance.\\

\section{Conclusion and Future Work}
\label{sec:conclusion}
This paper bridges the performance gap between state-of-the-art CNN-based graphical object detection algorithms and detection transformers. We perform different experiments on detection transformers and observe their effectiveness and efficiency for the graphical object detection task. The transformer-based approach can achieve higher table detection accuracy without relying on hand-crafted components like non-maximum suppression (NMS) and anchor design used in CNN-based object detectors. Furthermore, experimental results show that performance of detection tarnsformer depends on quality and quantity of object queries fed as input to transformer decoder. Consequently, these transformer-based detectors have achieved state-of-the-art performance on the four large public datasets, including TableBank, PubLayNet, PubTables and NTable dataset. In Future, we plan to extend the transformer-based model for table structure recognition and content extraction.

\noindent\textbf{Supplementary information}
The supplementary material contains a detailed explanation of the dataset,
evaluation setup and Implementation Details.

\section*{Declarations}
\noindent\textbf{Funding} The work has been partially funded by the European project INFINITY under Grant Agreement ID 883293.\par
\noindent\textbf{Conflict of interest} The authors declare that they have no conflict of interest.
\bibliography{main}% common bib file


\end{document}

% \title[Article Title]{Table Detection using Detection Transformer equipped with Query Denoising Training and Dynamic Anchor Boxes.}
\title[Supplemetary Material]{Supplemetary Material: Bridging the Performance Gap between DETR and R-CNN for Graphical Object Detection in Document Images.}

\newcommand{\orcidauthorA}{0000-0003-0456-6493} % Add \orcidA{} behind the author's name
\newcommand{\orcidauthorB}{0000-0002-0536-6867} % Add \orcidB{} behind
\newcommand{\orcidauthorC}{0000-0002-7052-979X}

\author*[1,2,3]{\fnm{Tahira} \sur{Shehzadi}}\email{tahira.shehzadi@dfki.de}

\author[1,2,3]{\fnm{Khurram Azeem} \sur{Hashmi}}\email{khurram\_azeem.hashmi@dfki.de}
\author[1,2,3]{\fnm{Didier} \sur{Stricker}}\email{didier.stricker@dfki.de}
\author[4]{\fnm{Marcus} \sur{Liwicki}}\email{marcus.liwicki@ltu.se}
\author[1,2,3]{\fnm{Muhammad Zeshan} \sur{Afzal}}\email{muhammad\_zeshan.afzal@dfki.uni-kl.de}

\affil[1]{\orgdiv{Department of Computer Science}, \orgname{Technical University of Kaiserslautern}, \city{Kaiserslautern}, \postcode{67663}, \country{Germany}}

\affil[2]{\orgdiv{Department of Computer Science}, \orgname{Mindgarage}, \city{Kaiserslautern}, \postcode{67663}, \country{Germany}}

\affil[3]{\orgdiv{Augmented Vision}, \orgname{German Research Institute for Artificial Intelligence (DFKI)},\city{Kaiserslautern}, \postcode{67663}, \country{Germany}}

\affil[4]{\orgdiv{Department of Computer Science}, \orgname{Luleå University of Technology)},\city{Luleå}, \postcode{97187}, \country{Sweden}}

\maketitle

The supplementary material contains a detailed explanation of the dataset, evaluation setup and Implementation Details. We explain these sections here because of space constraints. 

\section{Datasets and Evaluation Protocol}\label{secA1}
\subsection{Datasets}
\label{sec:datasets}
We perform experiments on three popular publicly available table detection datasets, including TableBank, PunLayNet and PubTables, to compare the effectiveness of anchor-based learning queries, denoising and improved denoising training modules on DETR performance. This section gives an overview of these datasets.

\noindent\textbf{TableBank}
\label{ghjgh}
The TableBank \cite{tablebank8} dataset is the second largest publicly accessible dataset for table detection tasks in the document analysis domain. The dataset contains 417K annotations of document images collected by applying the crawling process on the arXiv database. The dataset contains tables from three types of documents: having 163,417  word documents, 253,817 latex document and 417,234 Word + LaTeX documents. It also contains a table structure recognition dataset. For our experiment, we only consider the table detection dataset. \par

\noindent\textbf{PubLayNet}
PubLayNet \cite{PubLayNet3} is a large publicly available dataset containing 335,703 training set, 11,240 validation set and 11,405 test set document images. Its annotations include bounding boxes and polygonal segmentation of table, title, figure, text, and list of images taken from research papers and articles. This dataset used the coco analysis protocol for evaluation. For our experiment, we only take 102,514 table annotations of 86,460  images. \par

\noindent\textbf{NTable}
The NTable \cite{NTables} dataset is a collection of images designed for camera-based table detection and consists of three sub-datasets: NTable-ori, NTable-cam, and NTable-gen. NTable-ori contains original images of tables captured by cameras in various real-world settings, manually annotated with information about the location, size, and other attributes of the tables. NTable-cam is an augmented dataset created by applying various transformations to NTable-ori, and NTable-gen is a generated dataset synthesized using computer graphics techniques. The NTable dataset provides a comprehensive collection of images for training and testing camera-based table detection algorithms and can be used to develop and evaluate advanced methods for table detection and other applications in the future.

\noindent\textbf{PubTables}
The PubTables \cite{pubtables5} is the largest dataset having around one million tables taken from different scientific articles. The authors used the canonicalization method to resolve the over-segmentation issue noticed in previous datasets. It can be used for detecting tables, recognizing structure and functional analysis tasks as it has detailed information about location and header. For our experiment, we only consider table detection annotation of 947,642 tables. \par

\subsection{Evaluation Protocol}
\label{sec:metrics}

\noindent\textbf{Precision}
\label{sec:Precision}
Precision \cite{eval61} defines as the percentage of a predicted region that belongs to the ground truth. The formula for precision is explained below:

Precision is defined as the percentage of prediction area that fall within the ground-truth. The precision formula is:

\begin{equation}
\frac{\text{Predicted region in ground truth}} {\text{Total area of predicted region}}
 = \frac{\text{TP}}{\text{TP $+$ FP}}
\end{equation}

where TP stands for true positives and FP stands for false positives.

\noindent\textbf{Recall}
\label{sec:recall}
Recall \cite{eval61} is calculated as the percentage of ground truth region present in the predicted region. The recall formula is:

\begin{equation}
\frac{\text{Ground-truth region in predicted area}} {\text{Total area of ground truth region}}
 = \frac{\text{TP}}{\text{TP $+$ FN}}
 \end{equation}
 
TP stands for true positives and FN stands for false negatives. F1-Score
The F1-score \cite{eval61} is calculated as the harmonic-mean between the Precision and Recall.

\noindent\textbf{Average Precision}
\label{sec:ap}
In the context of MS COCO \cite{coco14} evaluation, Average Precision (AP) calculates the average precision at different recall levels. The higher the AP value, the better the performance and vice versa. The formula for average precision is expressed mathematically as:
\begin{equation}
\text{AP} = \sum_n (R_n - R_{n-1}) P_n    
\end{equation}
\par where R\textsubscript{n} and P\textsubscript{n} are the precision and recall at the n\textsubscript{th} threshold.

\noindent\textbf{Intersection Over Union}
\label{sec:IOU}
Intersection Over Union (IOU) \cite{iou} is one of the most important metrics regularly used to determine the performance of table recognition algorithms. This is a measure of how well the predicted region overlaps with the actual ground truth region. An IOU is defined as follows.
\begin{equation}
IOU = \frac{\text{Area of Overlap region}} {\text{Area of Union region}}
 \end{equation}

\noindent\textbf{Mean Average Precision}
\label{sec:map}
From MS COCO \cite{coco14} evaluation, Mean Average Precision (mAP) is another extensively applied evaluation metric for category-specific table detection. The mAP is the mean of average precision computed over all the classes. Mathematically, it is explained by:
Mean Average Precision (mAP) is another commonly used evaluation metric for category-specific object detection. mAP is the mean of the average precisions computed over all classes. Mathematically, it is explained as follows.
\begin{equation}
\text{mAP} = \frac{1}{N}\sum_{i=1}^{N} AP_i
\end{equation}
where $AP_i$ is the average precision for a particular class, described in section \ref{sec:ap}. $N$ is the total number of classes.

\subsection{Implementation Details}
All experiments are conducted with 8 Nvidia Tesla V100 GPUs. We used the ResNet-50 \cite{resnet45} backbone with the AdamW algorithm as an optimizer. We train the simple DETR model for 20 epochs, while with other modules for 12 epochs with a transformer's learning rate of 1e-4 and a backbone's learning rate of 1e-4. The weights of ResNet-50 are initialized with an ImageNet pre-trained model. The value of weight-decay is 1e-4. We set the number of layers value to 6 in the encoder-decoder module, and the value of the object query in the decoder input is 10. The object query value depends on the total number of classes of the dataset. For the coco dataset, its value is set to 100. However, for table detection datasets that have only one table class, set this value to 10.

\bibliography{sn-bibliography}% common bib file